\documentclass[preprint,12pt]{elsarticle}

\usepackage{csquotes}
\usepackage[commandnameprefix=ifneeded, final]{changes}

\usepackage{amssymb}
\usepackage[super]{nth}

\usepackage{float}

\usepackage{caption}
\usepackage{subcaption}

\usepackage{tabularray}

\usepackage{eurosym}

\usepackage{xurl}

\usepackage{hyperref}
\hypersetup{colorlinks=true,
linkcolor=blue,
filecolor=magenta,      
urlcolor=cyan,
pdfauthor=author}

\begin{document}
\journal{Knowledge-Based Systems}

\begin{frontmatter}
	
	
	
	\title{TreeC: a method to generate interpretable energy management systems using a metaheuristic algorithm.}
	
	
	\author[inst1]{Julian Ruddick\corref{cor1}}
	\ead{julian.jacques.ruddick@vub.be}
	\cortext[cor1]{Corresponding author}
	\affiliation[inst1]{organization={Electric Vehicle and Energy Research Group (EVERGI), Mobility, Logistics and Automotive Technology Research Centre (MOBI), Department of Electrical Engineering and Energy Technology, Vrije Universiteit Brussel},
				addressline={Pleinlaan 2}, 
				city={Brussels},
				postcode={1050},
				country={Belgium}}
	
	\author[inst1,inst2]{Luis Ramirez Camargo}
	\ead{l.e.ramirezcamargo@uu.nl}
	\author[inst1]{Muhammad Andy Putratama}
	\ead{Muhammad.Andy.Putratama@vub.be}
	\author[inst1]{Maarten Messagie}
	\ead{Maarten.Messagie@vub.be}
	\author[inst1]{Thierry Coosemans}
	\ead{Thierry.Coosemans@vub.be}

	\affiliation[inst2]{organization={Copernicus Institute of Sustainable Development -  Utrecht University},
				addressline={Princetonlaan 8a}, 
				city={Utrecht},
				postcode={3584 CB}, 
				state={Utrecht},
				country={Netherlands}}
	
	\begin{abstract}
	\replaced{\label{ch:abstract_begin}Energy management systems (EMS) have traditionally been implemented using rule-based control (RBC) and model predictive control (MPC) methods. 
	However, recent research has explored the use of reinforcement learning (RL) as a promising alternative. 
	This paper introduces TreeC, a machine learning method that utilizes the covariance matrix adaptation evolution strategy metaheuristic algorithm to generate an interpretable EMS modeled as a decision tree. 
	Unlike RBC and MPC approaches, TreeC learns the decision strategy of the EMS based on historical data, adapting the control model to the controlled energy grid.
	The decision strategy is represented as a decision tree, providing interpretability compared to RL methods that often rely on black-box models like neural networks. 
	TreeC is evaluated against MPC with perfect forecast and RL EMSs in two case studies taken from literature: an electric grid case and a household heating case.
	}{	
	Energy management systems (EMS) have classically been implemented based on rule-based control (RBC) and model predictive control (MPC) methods. 
	Recent research are investigating reinforcement learning (RL) as a new promising approach. 
	This paper introduces TreeC, a machine learning method that uses the metaheuristic algorithm covariance matrix adaptation evolution strategy (CMA-ES) to generate an interpretable EMS modeled as a decision tree. 
	This method learns the decision strategy of the EMS based on historical data contrary to RBC and MPC approaches that are typically considered as non adaptive solutions. 
	The decision strategy of the EMS is modeled as a decision tree and is thus interpretable contrary to RL which mainly uses black-box models (e.g. neural networks). 
	The TreeC method is compared to RBC, MPC and RL strategies in two study cases taken from literature: (1) an electric grid case and (2) a household heating case. 
	}
	\replaced{
	\label{ch:performance}
	In the electric grid case, TreeC achieves an average energy loss and constraint violation score of 19.2, which is close to MPC and RL EMSs that achieve scores of 14.4 and 16.2 respectively. 
	All three methods control the electric grid well especially when compared to the random EMS, which obtains an average score of 12\,875. 
	In the household heating case, TreeC performs similarly to MPC on the adjusted and averaged electricity cost and total discomfort (0.033 EUR/m\textsuperscript{2} and 0.42 Kh for TreeC compared to 0.037 EUR/m\textsuperscript{2} and 2.91 kH for MPC), while outperforming RL (0.266 EUR/m\textsuperscript{2} and 24.41 Kh). 
	}
	{The results show that TreeC obtains close performances than MPC with perfect forecast in both cases and obtains similar performances to RL in the electrical grid case and outperforms RL in the household heating case.}
	\replaced{TreeC demonstrates a performant and interpretable application of machine learning for EMSs.}{TreeC demonstrates a performant application of machine learning for energy management systems that is also fully interpretable.}
	
	\end{abstract}

	\begin{highlights}
		\label{ch:highlight}
	\item \added{TreeC generates performant and interpretable energy management systems.}
	\item \added{TreeC can perform better than model predictive control and reinforcement learning.}
	\item \added{TreeC's interpretability provides valuable insights on the controlled energy grid.}
	\item \added{A more complex energy management system model does not guarantee better performance.}
	\item \added{Reproducible benchmark cases are key to compare control methods.}
	\end{highlights}

	
	\end{frontmatter}

\section{Introduction}
\label{sec:sample1}

With the decreasing costs of renewable energy sources and energy storage \cite{bogdanov_low-cost_2021}, more energy is produced and stored in local energy grids without having to transit through a national grid. This also means that energy management systems (EMS) are necessary to manage these local energy grids and their assets \cite{olatomiwa_energy_2016}. EMSs have been deployed in wide-variety of scales and use cases of energy systems - from household levels with few assets, to nation-wide levels with various assets types and sizes. Most EMSs nowadays are implemented based on rule-based control (RBC) or model predictive control (MPC)\cite{olatomiwa_energy_2016,zia_microgrids_2018}. However these methods are deterministic and therefore do not adapt to the possible errors in forecasting or the differences between modeled and real grid behaviour. 
Research is done to make these methods adaptive \cite{bagwe_adaptive_2019,tesfay_adaptive-model_2018,herzog_self-adapting_2013,schmelas_savings_2017} as well as exploring different machine learning methods \cite{olatomiwa_energy_2016,zia_microgrids_2018}.

Reinforcement learning (RL) seems very well suited to the problem as it has mastered many complex games \cite{silver_mastering_2017}, adapts to the case it is implemented in and does not require a model of the controlled system contrary to the MPC approach \cite{wang_reinforcement_2020}. 
RL has been shown to work in simulation in heating, ventilation and air conditioning cases \cite{yu_multi-agent_2021,yang_towards_2021,gao_deepcomfort_2020}, in cases with multiple house appliances \cite{xu_multi-agent_2020,mocanu_-line_2019,li_real-time_2020,alfaverh_demand_2020}, in larger scale multi-energy systems \cite{ceusters_model-predictive_2021,yu_deep_2020} and in a distribution grid with PV production \cite{petrusev_reinforcement_2023}. Some of these studies obtain better results than MPC with imperfect forecasting \cite{yang_towards_2021,li_real-time_2020,yu_deep_2020, petrusev_reinforcement_2023} or even MPC with perfect forecast due to differences in the MPC and simulation models \cite{ceusters_model-predictive_2021}. It is difficult to reproduce or compare against these studies as in general the data and code used is not shared and some studies use commercial simulators \cite{gao_deepcomfort_2020} and data sets \cite{mocanu_-line_2019}.

To allow comparison between different EMS solutions, \citet{henry_gym-anm_2021} and \citet{arroyo_comparison_2022} have both published results comparing RL and MPC based EMS\added{s} with open access data and open source simulation environment. Moreover, Citylearn challenge, a yearly occurring EMS solution competition, also aims to establish benchmark cases with the first two editions requiring the participants to use RL based approaches \cite{vazquez-canteli_citylearn_2020,nagy_citylearn_2021}. 

There has been a push recently to use interpretable models for decision making when using machine learning techniques \cite{goodman_european_2017,rudin_stop_2019,barredo_arrieta_explainable_2020}. However the current research on RL based EMS\added{s} uses uninterpretable models (i.e. black box models) such as neural networks. The main reason to use a black box model would be if it has a significant performance advantage over the interpretable models. Making decision modes more interpretable would provide multiple benefits such as having total confidence in the deployed model, being able to interpret either in advance or retrospectively why an EMS could not work and makes it easier to transfer the knowledge gained on how to control one system to another. 

Decision trees are interpretable models appreciated in the machine learning due to their good performances in many applications \cite{barredo_arrieta_explainable_2020,quinlan_induction_1986,jordan_machine_2015}.

\citet{huo_decision_2021} uses decision trees based EMS\added{s} for an energy grid application. 
The method first generates a good policy with an MPC and then generates the decision trees to reproduce the MPC behaviour by using classification. 
\citet{dai_deciphering_2023} uses a RL based EMS with a neural network model then generates a decision tree to explain the behaviour of the neural network using classification. 
Both these papers produce interpretable or explainable EMS\added{s} but in an indirect way where the optimised \replaced{EMS strategy is approximated to decision trees using classification.}{actions of the EMS are generated by another method then this behaviour is approximated using decision trees.}

Generating interpretable and black box control models using a metaheuristic algorithms is common practice in fields such as robotics \cite{urzelai_incremental_1998,francesca_automode_2014,ligot_automatic_2020}, power systems \cite{ibraheem_recent_2005} and also EMS \cite{shaikh_review_2014}.
Generating decision trees specifically has also been done to solve control problems \cite{kretowski_memetic_2008,custode_evolutionary_2023} but to the best of the author's knowledge not for EMS\added{s}.

A metaheuristic based approach generally generates a model in simulation before deploying it in real-life. 
The advantage of this approach is being able to do many simulations to find a good model. 
However it exposes the generated model to a drop in performance due to the difference between simulation and real life \cite{jakobi_noise_1995} and requires to have a simulation of the case. 
RL on the other hand can be implemented directly in real-life without a need of a simulation but the performance is generally very poor for the first steps meaning some additional transfer learning is necessary to improve the performance, a problem to which no satisfying solution has been found yet \cite{wang_reinforcement_2020}.

This paper presents TreeC, a machine learning method based on the metaheuristic algorithm covariance matrix adaptation evolution strategy (CMA-ES). 
TreeC generates an EMS modeled as interpretable decision trees. 
The performance of this method is evaluated against the EMS\added{s} presented in \cite{henry_gym-anm_2021} and \cite{arroyo_comparison_2022} which both made available the code and data necessary to perform a comparison with the RL and MPC based methods they have implemented. 

\added{
	\label{ch:novelty_intro}To the best of the author's knowledge, TreeC is the first method to optimise the generation of an EMS modeled as decision trees directly on its simulation performance instead of approximating an EMS strategy generated through another method like in \citet{huo_decision_2021} and \citet{dai_deciphering_2023}.
	}

\section{Method}
\label{sec:method}

\subsection{TreeC generation method}

\begin{figure}[H]
     \centering
     \begin{subfigure}[b]{0.9\textwidth}
         \centering
         \includegraphics[width=\textwidth]{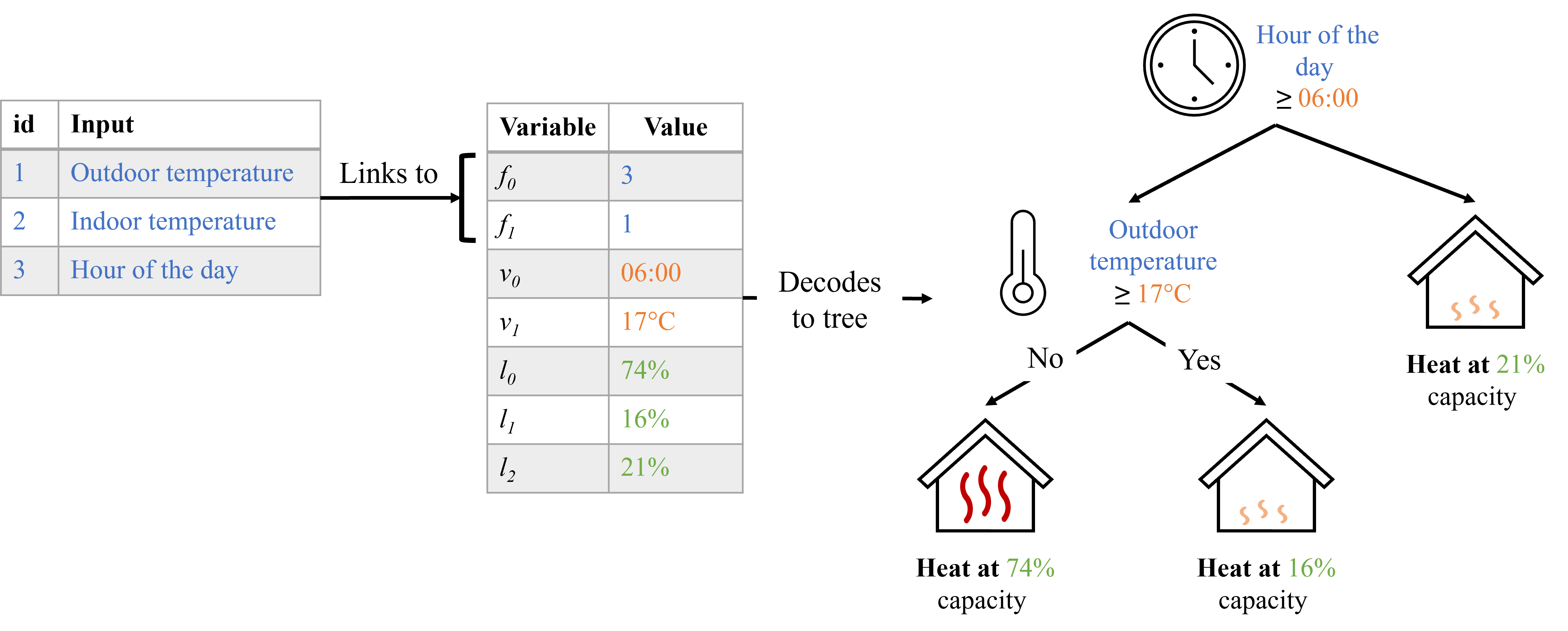}
         \caption{Encoding}
         \label{fig:encoding}
     \end{subfigure}
     \hfill
     \begin{subfigure}[b]{0.9\textwidth}
         \centering
         \includegraphics[width=\textwidth]{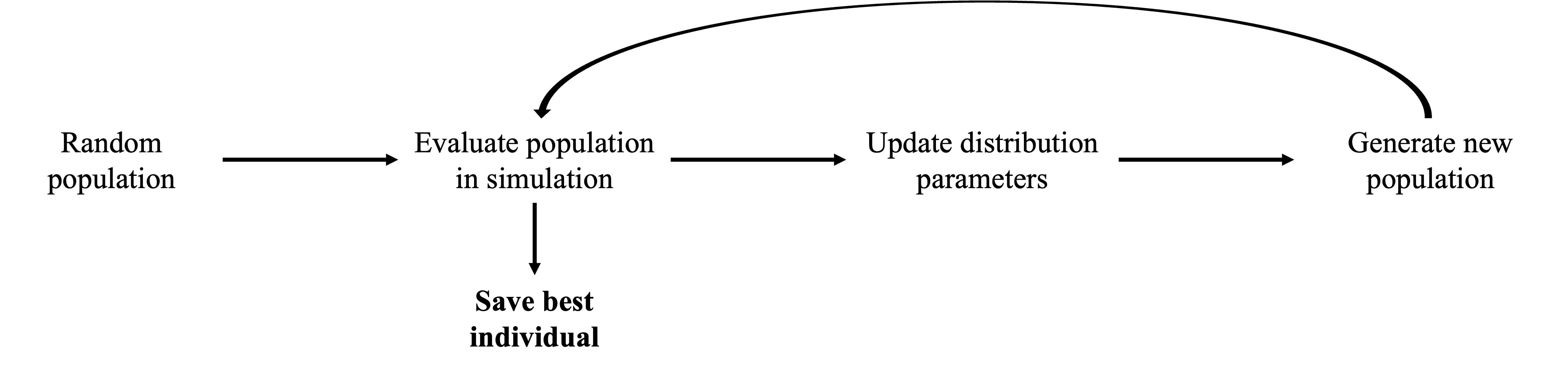}
         \caption{Optimisation}
         \label{fig:optimisation_proc}
        \end{subfigure}
     \hfill
     \begin{subfigure}[b]{0.85\textwidth}
         \centering
         \includegraphics[width=\textwidth]{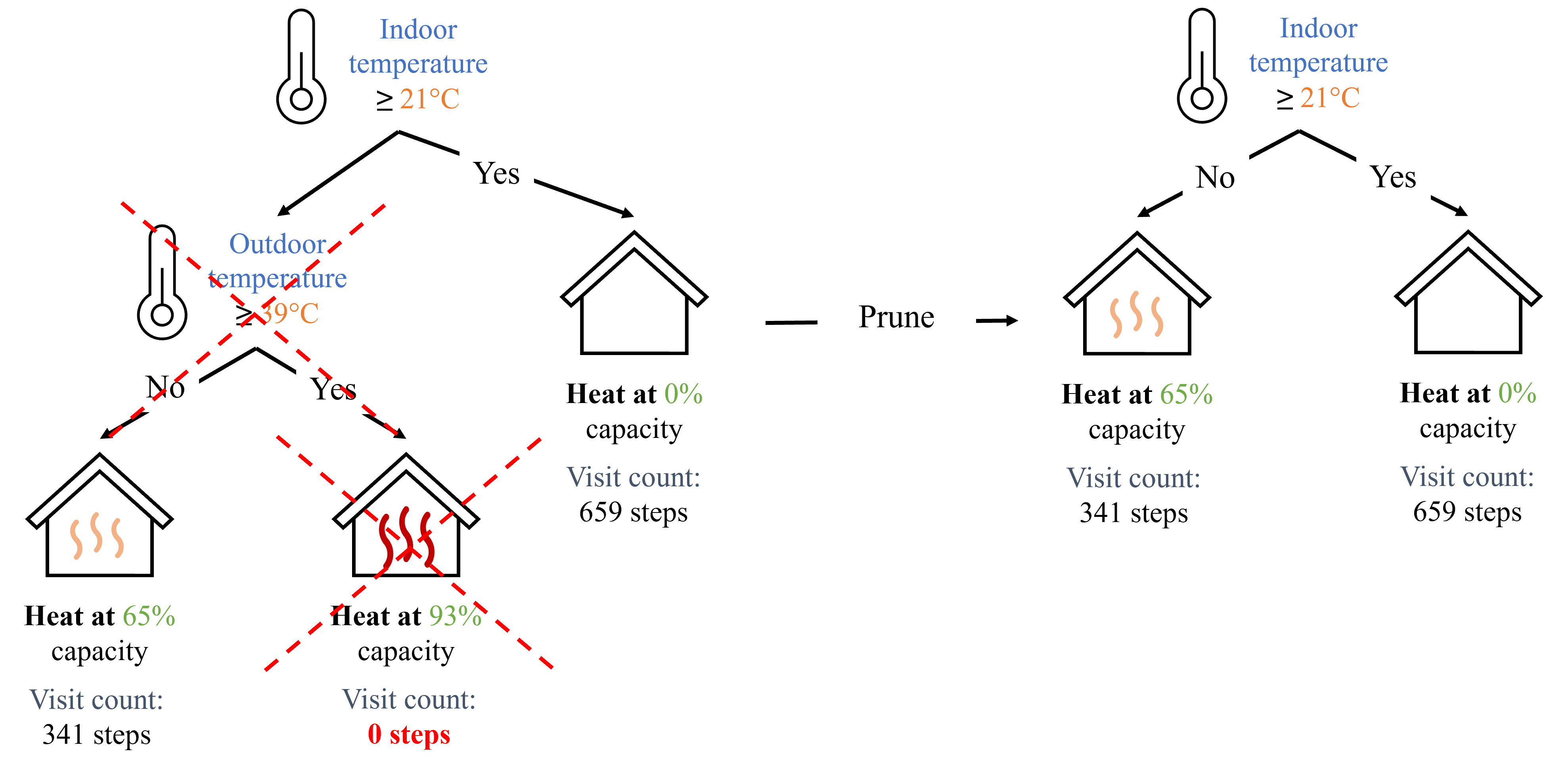}
         \caption{Pruning}
         \label{fig:pruning}
     \end{subfigure}
        \caption{Illustration of TreeC for a household heating case. Fig. \ref{fig:encoding} shows an decision tree energy management system and how it is encoded as a list of numbers, Fig. \ref{fig:optimisation_proc} shows the optimisation process of the covariance matrix adaptation evolution strategy algorithm and Fig. \ref{fig:pruning} shows an example of how a tree is pruned when a leaf node is not visited.}
        \label{fig:method_viz}
\end{figure}

TreeC has three important operations: 
\begin{itemize}
   
\item Encoding (Fig. \ref{fig:encoding}): to translate a list of numbers to the tree EMS.
 \item Optimisation (Fig. \ref{fig:optimisation_proc}): to successively improve the trees.
 \item Pruning (see Fig. \ref{fig:pruning}): to reduce the size of the tree by removing unused leafs and nodes.
\end{itemize}

The decision tree used is a complete binary tree. 
The tree is encoded on $3N+1$ variables with $N$ number of parent or split nodes and $N+1$ number of leafs. 
The first $2N$ variables represent the features ($f_0, f_1,...,f_N$) and the values ($v_0,v_1,...,v_N$) of the node split. 
The remaining $N+1$ variables ($l_0,l_1,...,l_{N+1}$) are the values that represent which actions should be taken on each leaf as illustrated in Fig.\ref{fig:encoding}. 
\added{\label{ch:var_explanation}The $3N+1$ variables defining the tree EMS are the optimisation variables of the energy management problem and are modified to optimise the performance of the EMS on this problem.}  
This \added{encoding} approach is commonly used in many applications with different variations as shown in the review by \citet{barros_survey_2012}.

Note that the number of parent nodes $N$ is an important hyperparameter that influences the complexity of the behaviour and how interpretable the decision tree is. In this method, the $N$ value is set to 20 as this allows a relatively complex behaviour while keeping the tree interpretable after pruning.

\label{ch:explanation_cmaes}The proposed method uses the state-of-the-art metaheuristic algorithm CMA-ES \cite{hansen_reducing_2003}. 
CMA-ES was chosen due to its high performance in different optimisation problems \cite{hansen_comparing_2010} and it having almost no user-defined hyperparameters. 
In CMA-ES, a set of random individuals are generated for the initial population, these individuals are evaluated by an objective function. 
The score obtained by this evaluation then modifies the distribution parameters used to generate the individuals of the next population. 
This process is repeated multiple times in order to obtain individuals with a better score over the generations. 
The final selected EMS is the individual that obtained the best score over the whole optimisation process  (see Fig. \ref{fig:optimisation_proc}). 

The default CMA-ES population size of $4+3\lfloor ln(n)\rfloor $ is used here with $n$ being the number of variables defining the decision tree \cite{hansen_principled_2014}. 

It is important that the score obtained by each individual are comparable meaning that the evaluation conditions should be exactly the same (i.e. same start time, evaluation period, ...). 

Finally the decision trees generated have leafs that were never reached during the training phase and therefore should be pruned.The pruning methods checks if each leaf has been reached and if not removes the leaf and replaces the parent node by it's other child node. This process is repeated until all remaining leaves have been visited at least once (see Fig.\ref{fig:pruning}).

\subsection{Parallel training and multi-threading properties}
\label{sec:par_train}

Metaheuristics algorithms are good to solve complex problems but are non-deterministic so two separate trainings will not find the same solution. To obtain a better solution, five different trainings are executed with the same problem conditions but different initial population for the CMA-ES algorithm. The tree with the best score in training is then evaluated to obtain the results. Doing this also allows to observe if the trees stemming from different trainings have recognisable similar behaviours or which are the most used features. 

This method can be executed using many parallel threads. 
The CMA-ES algorithm allows the evaluation of the individuals within a population to be done in parallel. 
This means the maximum number of possible threads is equal to the number of individuals within a population used for the training multiplied by the number of different parallel training (five in this case).

\begin{figure}[H]
	\centering
	\includegraphics[width=1.0\linewidth]{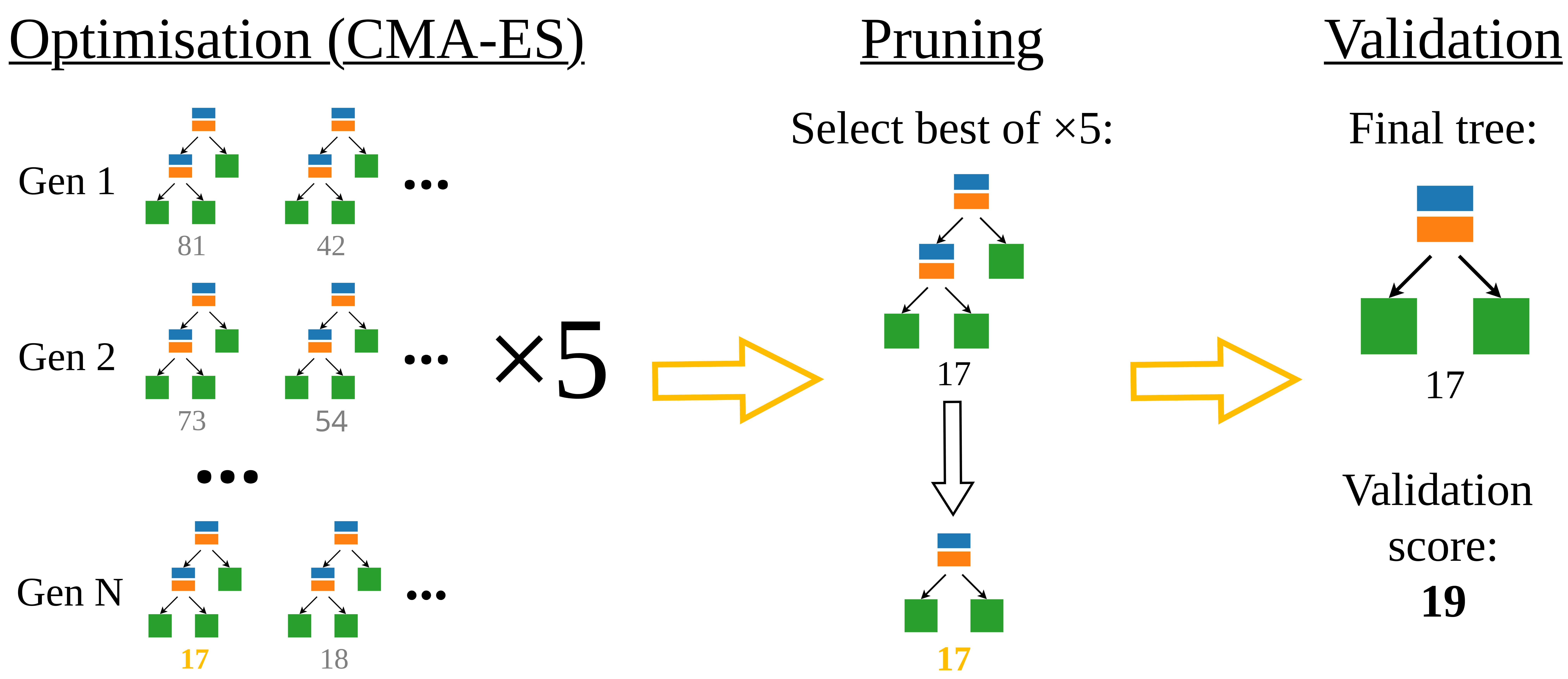}
	\caption{\added{Summary of the optimisation/training, pruning and validation method. 
	The decision trees represent different evaluated candidates with the score they obtained beneath them and uses the same color code for variables as in Fig. \ref{fig:method_viz}.
	The decision trees with yellow scores are the best performing ones and taken to the next step.
	The optimisation is executed 5 times in parallel as described in Section \ref{sec:par_train}.}}
	\label{fig:summary_method}
\end{figure}

\added{Fig. \ref{fig:summary_method} shows a summary of the training, pruning and validation process including the five parallel training described in this section.}

\subsection{Case study 1: ANM6easy case - electrical grid}

\subsubsection{Case description}

\begin{figure}[H]
\centering
\includegraphics[width=1.0\linewidth]{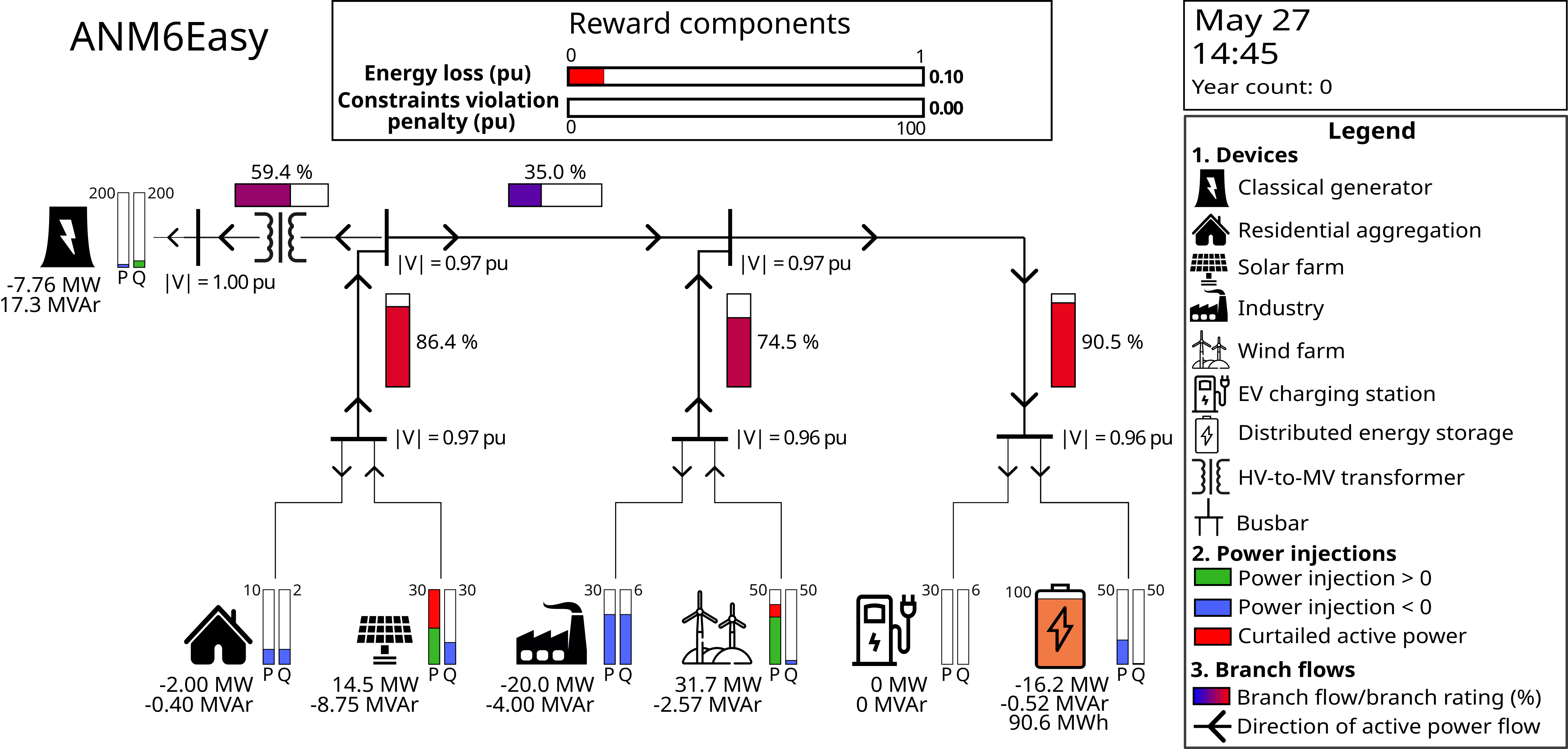}
\caption{ANM6easy case visualisation for a typical working hour period using the EMS presented in Fig. \ref{fig:anm_trees}. All the assets are shown with their respective real and reactive power as well as the constraints on the grid lines and voltage magnitude of the busbars \cite{henry_gym-anm_2021-1}. }
\label{fig:ANM case}
\end{figure}

The first case study is ANM6easy a test case proposed  by \cite{henry_gym-anm_2021} and provided in the gym-anm simulator. The version 1.0.2. of gym-anm was used in this work. ANM6Easy simulates an electrical grid as shown in Fig. \ref{fig:ANM case} It is composed of three passive loads (an industrial complex, a residential area and an electric vehicle charging garage), a slack fossil fuel generator, and controllable assets that consist of two renewable energy sources (i.e., a wind-turbine farm and a photovoltaic (PV) plant) and one large electrical storage unit. The renewable energy sources and the large electrical storage unit are controllable. For the renewable assets, the EMS decides the curtailment of real power and the reactive power. For the electrical storage unit, the EMS decides the real and reactive power set-point. The decisions are taken by the EMS for the next 15 minutes. 
The simulation repeats the same day over and over in which there are three distinct periods.

\begin{itemize}
    \item From 23:00 to 06:00: Windy night with very low demand from the three passive loads.
    \item From 08:00 to 11:00 and from 18:00 to 21:00: Typical arrival time at work or home. Very high demand from the electric vehicle charging garage due to people plugging in their vehicles to charge. A low amount of power is produced by both renewable asset plants. The demand from the residential area is high while for the industrial complex it is moderate.
    \item From 13:00 to 16:00: Typical working hours. The demand for the industrial area is high while the residential area demand is low. Both renewable asset plants produce a lot of power. There is no power demand from the electric vehicle charging garage. 
\end{itemize}

During the hours not included in the three characteristic periods, the non-controllable powers shift linearly from the powers of the preceding period to the powers of the next period.

The electrical grid can collapse if mismanaged, most often this occurs due to a voltage collapse problem \cite{chiang_voltage_1990}.

The objective of the EMS in this test case is to control the real and reactive powers of the renewable asset plants and large electrical storage unit in order to: (1) avoid exceeding the maximum power allowed on the branches, (2) maintain the voltage within nominal range with a tolerance of $\pm 5\%$ and (3) minimise energy loss. 

\added{The performance metric for this case depends on the reward $r_t$ calculated using Eq. (\ref{eq:reward_ANM}):}

\begin{equation}
	\added{r_{t}=clip(-100,\Delta E_{t:t+1}+1000\phi (s_{t+1}),100)}
	\label{eq:reward_ANM}
\end{equation}

\added{where $\Delta E_{t:t+1}$ is the total energy loss between time-step $t$ and $t+1$ and $\phi (s_{t+1})$ is a penalty term associated with the violation of the constraints of the grid. 
The penalty term $\phi (s_{t+1})$ increases when the maximum power of a branch is exceeded and when a bus voltage magnitude is not kept within the 0.95 to 1.05 per unit range. 
Both $\Delta E_{t:t+1}$ and $\phi (s_{t+1})$ are expressed in per unit with the base power of the system set to 100 MVA. When the electrical grid collapses, $r_t$ is set to 20\,000.
}

\added{The energy loss and constraint violation score $S$ is the same as defined in the original paper \cite{henry_gym-anm_2021}. 
It is calculated over 3\,000 time-steps and uses an additional discount factor $0.995^t$ (see Eq. (\ref{eq:obj_valid_anm})). 
}

\begin{equation}
	\added{S=\sum_{t}^{3000}r_t*0.995^t}
\label{eq:obj_valid_anm}
\end{equation}

\subsubsection{Training and validation}

During the training, six trees are evolved simultaneously. 
A different tree is used for the reactive power and real power of each of the three controllable assets. 
There are 18 possible split features to select from. 
These features are\added{:} the real and reactive power \replaced{of}{for} each asset \added{\label{ch:cascading_comment} from the previous time-step} (14 features), the real power of the renewable asset plants \added{from the previous time-step} if they were not curtailed (two features), the \added{current} state of charge of the large electrical storage unit and the \added{current} time of the day. 
The values of the leafs sets the power for the next time-step. 

For the results, 20 sets of six trees are generated using the method described in Section \ref{sec:method}. The problem has 366 variables, a population of 21 is used and the training was done over 1500 generations. During training, the population's individual are evaluated over a period of 300 time-steps (~3 days). The start time of this time-period is defined by a random seed and a different random seed is assigned to each generation process of the final 20 sets of six trees. 

The objective function of the training (Fig. \ref{fig:optimisation_proc}) and validation both use the reward $r_t$ provided by the simulator. Eq. (\ref{eq:reward_ANM}) shows how $r_t$ is calculated:

\deleted{$r_{t}=clip(-100,\Delta E_{t:t+1}+1000\phi (s_{t+1}),100)$}

\deleted{where $\Delta E_{t:t+1}$ is the total energy loss between time-step $t$ and $t+1$ and $\phi (s_{t+1})$ is a penalty term associated with the violation of the constraints of the grid. 
The penalty term $\phi (s_{t+1})$ increases when the maximum power of a branch is exceeded and when a bus voltage magnitude is not kept within the 0.95 to 1.05 per unit range. 
Both $\Delta E_{t:t+1}$ and $\phi (s_{t+1})$ are both expressed in per unit with the base power of the system set to  100 MVA. 
When the electrical grid collapses, $r_t$ is set to 20\,000.
}

The objective function for training $O_{train}$ is a sum of the rewards $r_t$ (see Eq. (\ref{eq:reward_ANM})) for 300 time-steps (see Eq. (\ref{eq:obj_train_anm})). 

\begin{equation}
O_{train}=\sum_{t}^{300}r_t
\label{eq:obj_train_anm}
\end{equation}

\deleted{The objective function for validation $O_V$ is the same as defined in the original paper \cite{henry_gym-anm_2021}. 
It is calculated over 3\,000 time-steps and uses an additional discount factor $0.995^t$ (see Eq. (\ref{eq:obj_valid_anm})). }

\deleted{$O_V=\sum_{t}^{3000}r_t*0.995^t$}

The discount factor \added{used in the energy loss and constraint score $S$ (see Eq. (\ref{eq:obj_valid_anm}))} was removed for the training objective function $O_{train}$ because it favors earlier time-steps of the simulation\deleted{ and since the start-time of the validation evaluation is different from the start-time of the training evaluation}. 
\replaced{Removing the discount factor makes}{By removing the discount factor, the aim is to make} the reward of all time-steps equally important.

The \replaced{validation}{final} results show \added{for each generated EMS} the average \replaced{energy loss and constraint score $S$}{score obtained using the validation objective function} (see Eq. (\ref{eq:obj_valid_anm})) \added{calculated} over 10 different random start times (seeds 0 to 9 in gym-anm simulator). 
The results for the benchmark EMSs MPC with constant forecast, MPC with perfect forecast, Proximal Policy Optimization (PPO) \replaced{,}{and} Soft Actor-Critic (SAC)\added{ and random} were recalculated using the method of the original paper \cite{henry_gym-anm_2021} and evaluated using the metric described in the previous sentence.

\subsection{Case study 2: BOPTEST Hydronic Heat Pump case - household heating}

\subsubsection{Case description}

The second case study is BOPTEST Hydronic Heat Pump a test case provided in the BOPTEST simulator \cite{blum_prototyping_2019}. 
The version 0.1.0. of BOPTEST was used in this work. 
The case is a heating management problem in a household. 
In particular, this test case consist of a five person house with a rectangular floor of 12 by 16 meters and a height of 2.7 meters. 
The house is equipped with a water based floor heating, the water is heated by an air-to-water modulating heat pump of 15 kW nominal heating capacity. 
The interior temperature is impacted by the outdoor weather conditions. 
The walls, floor, roof and windows have thermal properties described extensively in the documentation of the test case \cite{blum_documentation_2021}. 
The simulation uses one year of weather data from Brussels provided by the simulator \cite{blum_prototyping_2019}. 

Three different energy pricing scenarios are made available by the simulator:

\begin{itemize}
    \item The constant electricity price: Fixed electricity \replaced{price}{cost} of 0.2535 \euro/kWh.
    \item The dynamic electricity price: During on-peak hours between 07:00 and 22:00 the electricity \replaced{price}{cost} is  0.2666 \euro/kWh and between 22:00 and 07:00 the electricity \replaced{price}{cost} is 0.2383 \euro/kWh.
    \item The highly dynamic price: Hourly dynamic electricity price taken from the Belgian day-ahead energy prices of 2019. The median electricity price is 0.2389 \euro/kWh with a lowest quartile price of 0.2317 \euro/kWh and an upper quartile price of 0.2392 \euro/kWh.
\end{itemize}

The simulator also provides two validation periods of 14 days each. The first spans from the \nth{17} to the \nth{31} of January and is centered around the peak heat day of the year, the second period spans from the \nth{19} April to the \nth{3} May and is centered around the typical heat day of the year.

The house is considered occupied before 7:00 and after 20:00 every weekday and at all times during the weekend. The comfort range is between 21°C and 24°C during occupied hours and between 15°C and 30°C during unoccupied hours. 

The objective of the case is to control the heat pump modulation signal for compressor frequency in order to keep the temperature of the house within the comfort range when occupied and minimise the electricity costs.

\added{The two performance metrics for this case are the total electricity cost and the total discomfort.
}

\added{
The total electricity cost $E$ is calculated using Eq. (\ref{eq:elec_cost_boptest}):
}

\begin{equation}
\added{E=\sum_{t}^{T}p_t*e_t}
\label{eq:elec_cost_boptest}
\end{equation}

\added{where $T$ is the set of simulated time-steps, $p_t$ is the price of electricity and $e_t$ is the electrical consumption of the heating system composed of the heat pump, the evaporator fan and the circulation pump.
}

\added{
The total discomfort $D$ is calculated using Eq. (\ref{eq:discomfort_boptest}):
}

\begin{equation}
\added{D=\sum_{t}^{T}\delta_t}
\label{eq:discomfort_boptest}
\end{equation}

\added{where $T$ is the set of simulated time-steps and $\delta_t$ is the discomfort metric.
$\delta_t$ has two different possible values: (1) when the indoor house temperature is out of the comfort range it equals the absolute difference between the indoor house temperature and the closest comfort temperature range bound, (2) when the indoor house temperature is within the comfort range, $\delta_t=0$.}

\deleted{The compared results in Fig. \ref{fig:boptest_tree_res} were taken from the original paper \cite{arroyo_comparison_2022}.}

\subsubsection{Training and validation}
\label{sec:train_val_boptest}
For this case, only one tree is necessary to control the heat pump. 
The tree has the same inputs and outputs as the SS2 RL based EMS presented in \cite{arroyo_comparison_2022}. 
There are 5 possible split features which are the price of electricity, the indoor house temperature, the lower and upper comfort range bounds and the time of the week. 
The values of the leafs sets the heat pump modulation signal for compressor frequency for the next time-step. 
They are discretised to 11 values between 0 and 1 with a step of 0.1.

Two different trees are generated for each pricing scenario and validation period. The problem has 61 variables, a population of 16 is used and the training was done over 150 generations. The evaluation of the individuals was done over 14 days starting 15 days before the start of the validation period and with a warm-up period of 1 day performed by the simulator. The heat pump is controlled every 15 minutes.

The objective function $O_{train}$ used for training is the same used by \citet{arroyo_comparison_2022} to obtain their results also shown in this paper. 
A weighted objective function is used to take into account the discomfort and \replaced{electricity}{operational} cost aspects of the case. 
The following Eq. (\ref{eq:obj_train_boptest}) shows how $O_{train}$ is calculated:

\begin{equation}
\added{O_{train}=w_1*D+w_2*E}
\label{eq:obj_train_boptest}
\end{equation}
\deleted{$O_{train}=\sum_{t}^{T}w_1*\delta_t+w_2*(p_t*e_t)$}

where \deleted{$T$ is the set of simulated time-steps, }$w_1$ and $w_2$ are the weights given to the \added{total }discomfort \added{$D$ }and \replaced{total electricity}{operational} cost \replaced{$E$ respectively}{metrics}.
\replaced{The weights}{They} are set to 100 and 192 respectively to match the weights used in \cite{arroyo_comparison_2022}. 
\deleted{The discomfort metric $\delta_t$ has two different possible values: (1) when the indoor house temperature is out of the comfort range it equals the absolute difference between the indoor house temperature and the closest comfort temperature range bound, (2) when the indoor house temperature is within the comfort range it, 
$\delta_t=0$. $p_t$ is the price of electricity and $e_t$ is the electrical consumption of the heating system composed of the heat pump, the evaporator fan and the circulation pump.}

\added{The validation results show the total electricity cost $E$ and total discomfort $D$ obtained for the generated TreeC EMSs.
 These results are compared to the performance of other EMSs extracted from the original paper \cite{arroyo_comparison_2022}.}

\section{Results}

The results and code necessary to reproduce the following results are available on the public treec-paper-results GitHub repository\footnote{\url{https://github.com/EVERGi/treec-paper-results}}. The training for the results of the two case studies were obtained on a high performance computer composed of 880 Intel\textregistered Xeon\textregistered Gold 6148 processors.

\subsection{ANM6easy case}

\begin{figure}[H]
\centering
\includegraphics[width=0.75\linewidth]{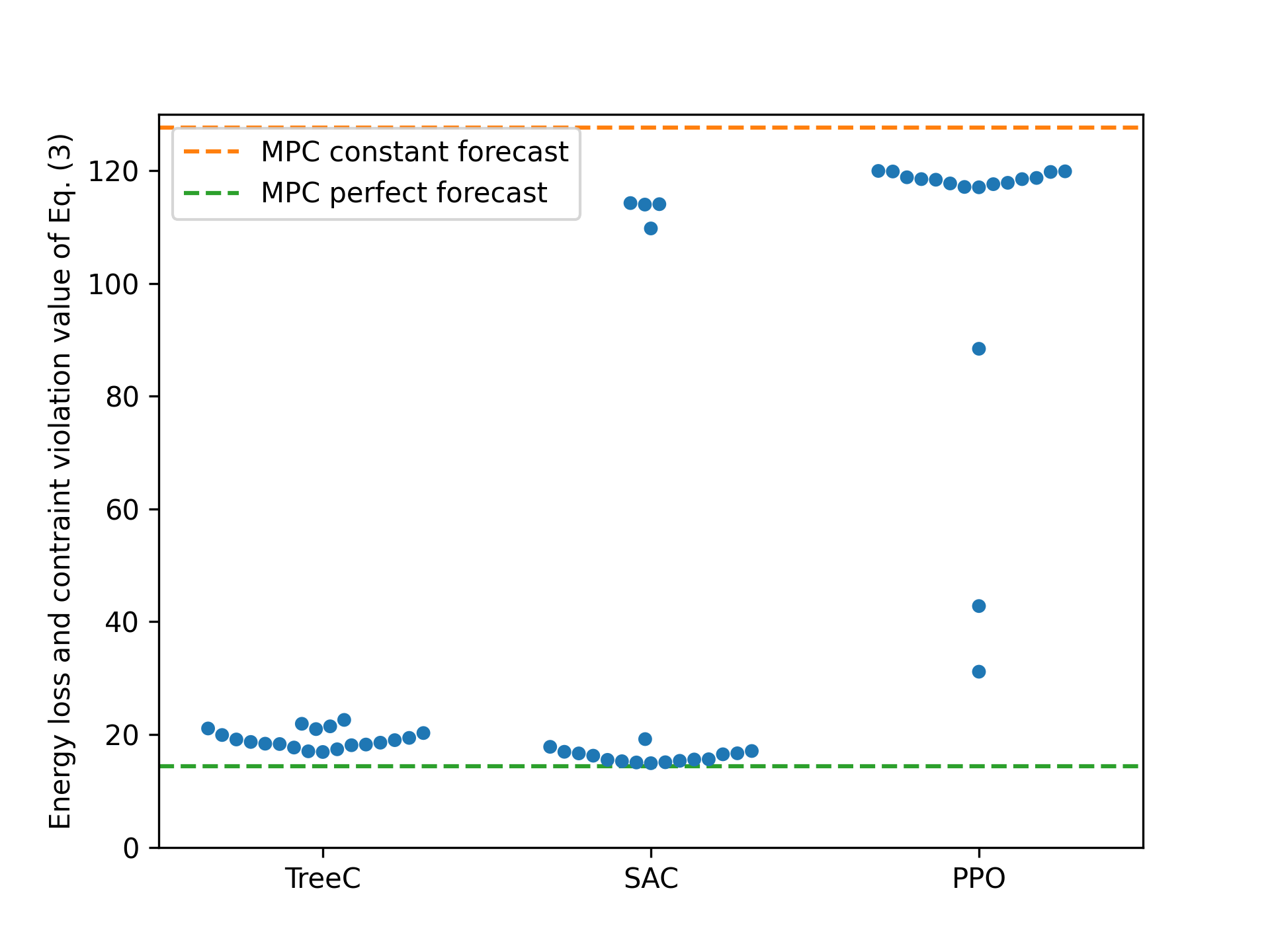}
\caption{Swarm plot representation of the results obtained for the ANM6easy case with each point being the validation of independant EMSs and the horizontal lines being the validation of the MPC EMSs. The TreeC and SAC EMSs both obtain results close to the MPC with perfect forecast. Disregarding its outliers, SAC performs slightly better than TreeC. PPO performs worse than the two other methods. Three outliers for PPO with a score higher than 130 are not shown in the plot.}
\label{fig:ANM_all_box}
\end{figure}

In the first study case, running the evaluation of one individual simulation of 300 time\replaced{-}{ }steps lasted ~10 seconds. The time to update the CMA-ES optimisation model is negligible here. The trees generated for the results shown in Fig. \ref{fig:ANM_all_box} all ran for  157\,500 evaluation which amounts to ~18 days 5 hours and 30 minutes of training using a single thread and to ~4 hours and 10 minutes using the maximum number of 105 threads (population size of 21 $\times$ 5 parallel training).

In this case, the TreeC EMS perform much better than PPO and MPC with constant forecast methods and slightly worse than the MPC with perfect forecast and SAC method (see Fig. \ref{fig:ANM_all_box}). 
The average score of SAC method is worse than TreeC due to outliers but it would be unfair to claim that TreeC performs better since it selects the best tree form five trainings to avoid such outliers. 
Implementing such a mechanism in the SAC method would improve its average score greatly. 

All methods perform much better than \replaced{the random}{an} EMS \replaced{that samples}{sampling} random actions from the possible action range of each controllable asset. 
\replaced{This}{The} random EMS obtains an average score of \replaced{12\,875 over 200 runs}{13\,348 over a 100 runs (score calculated with Eq. (\ref{eq:obj_valid_anm}))}. 
The electrical grid collapsed in each of the \replaced{200}{100} runs after \replaced{181}{159} steps on average.
All EMS methods presented in Fig. \ref{fig:ANM_all_box} avoid collapses consistently. 

\begin{table}[H]
	\centering
	\caption{\added{Average and standard deviation of the EMSs scores for the ANM6easy case.}}

\begin{tabular}{|c|c|}

	\hline
	
	\added{EMS} & \added{Energy loss and constraint violation score}  \\ \hline
	\added{MPC perfect forecast} & \added{\textbf{14.4 $\pm$ 0}}\\ \hline
	\added{SAC no outlier} & \added{16.2 $\pm$ 1.2}\\ \hline
	\added{TreeC} & \added{19.2 $\pm$ 1.7}\\ \hline
	\added{SAC} & \added{35.6 $\pm$ 39.7} \\ \hline
	\added{PPO} & \added{120.9 $\pm$ 44.9}  \\ \hline
	\added{MPC constant forecast} & \added{127.7 $\pm$ 0}\\ \hline
	\added{Random EMS} & \added{12\,875 $\pm$ 1308.8}\\ \hline 

	\end{tabular}
	\label{tab:ANM_table_results}
	
\end{table}

\added{Tab. \ref{tab:ANM_table_results} provides the average score obtained by each EMS method from best to worst.
The table also includes the performance of SAC without the four outliers that obtained a score higher than 100. 
The abstract, Tab. \ref{tab:comp} and conclusion use the average SAC score without outliers as the reference score for RL on the ANM6easy case.
We believe this is a more fair comparison because the TreeC method selects the best tree from five trainings to avoid such outliers (see Section \ref{sec:par_train}).
Such a mechanism was not implemented in the SAC method to keep the exact same implementation as the original paper \cite{henry_gym-anm_2021}.
}

\begin{figure}[H]
     \centering
     \begin{subfigure}[b]{0.45\textwidth}
         \centering
         \includegraphics[width=\textwidth]{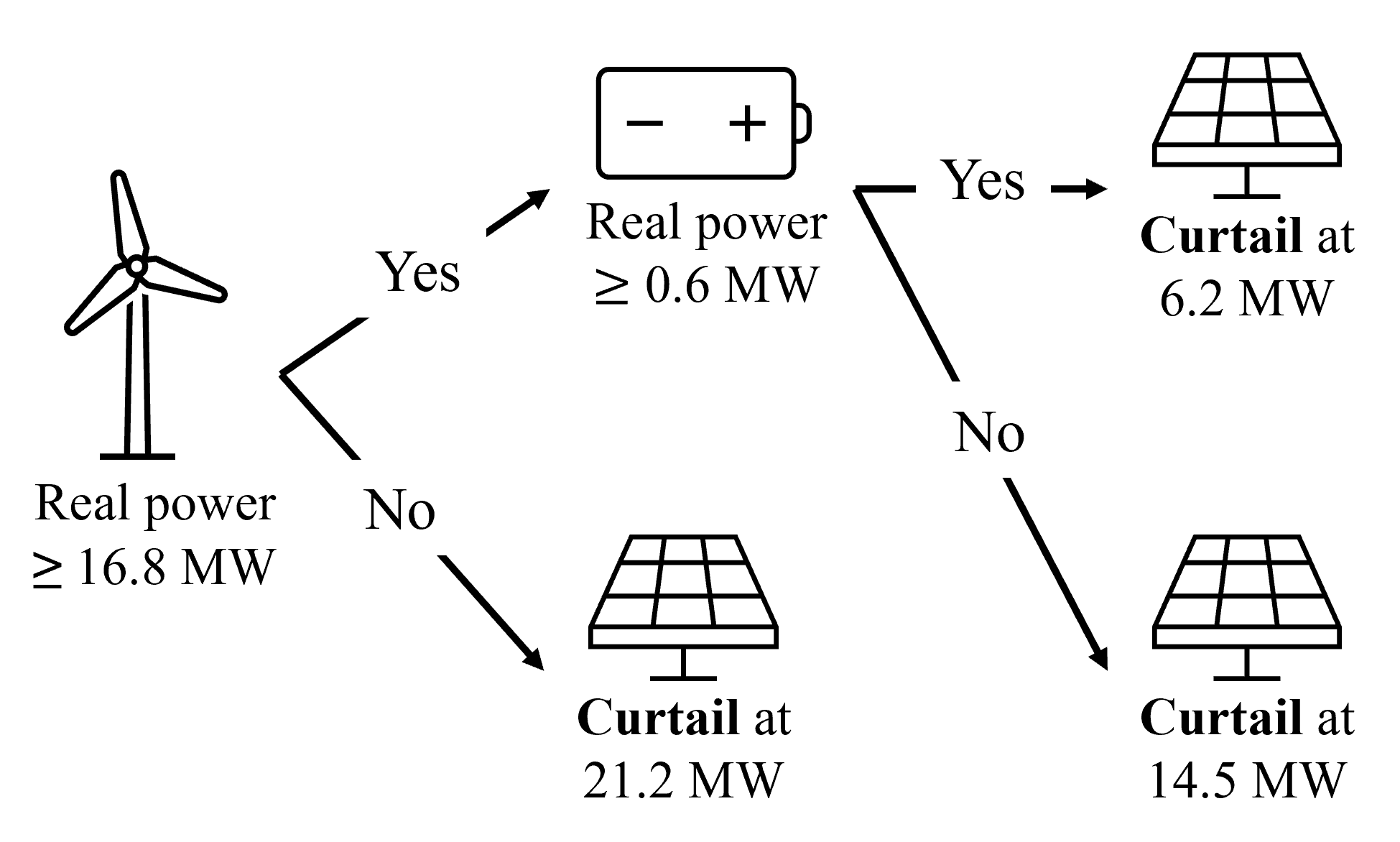}
         \caption{Curtailment photovoltaics}
         \label{fig:curt_pv}
     \end{subfigure}
     \hfill
     \begin{subfigure}[b]{0.45\textwidth}
         \centering
         \includegraphics[width=\textwidth]{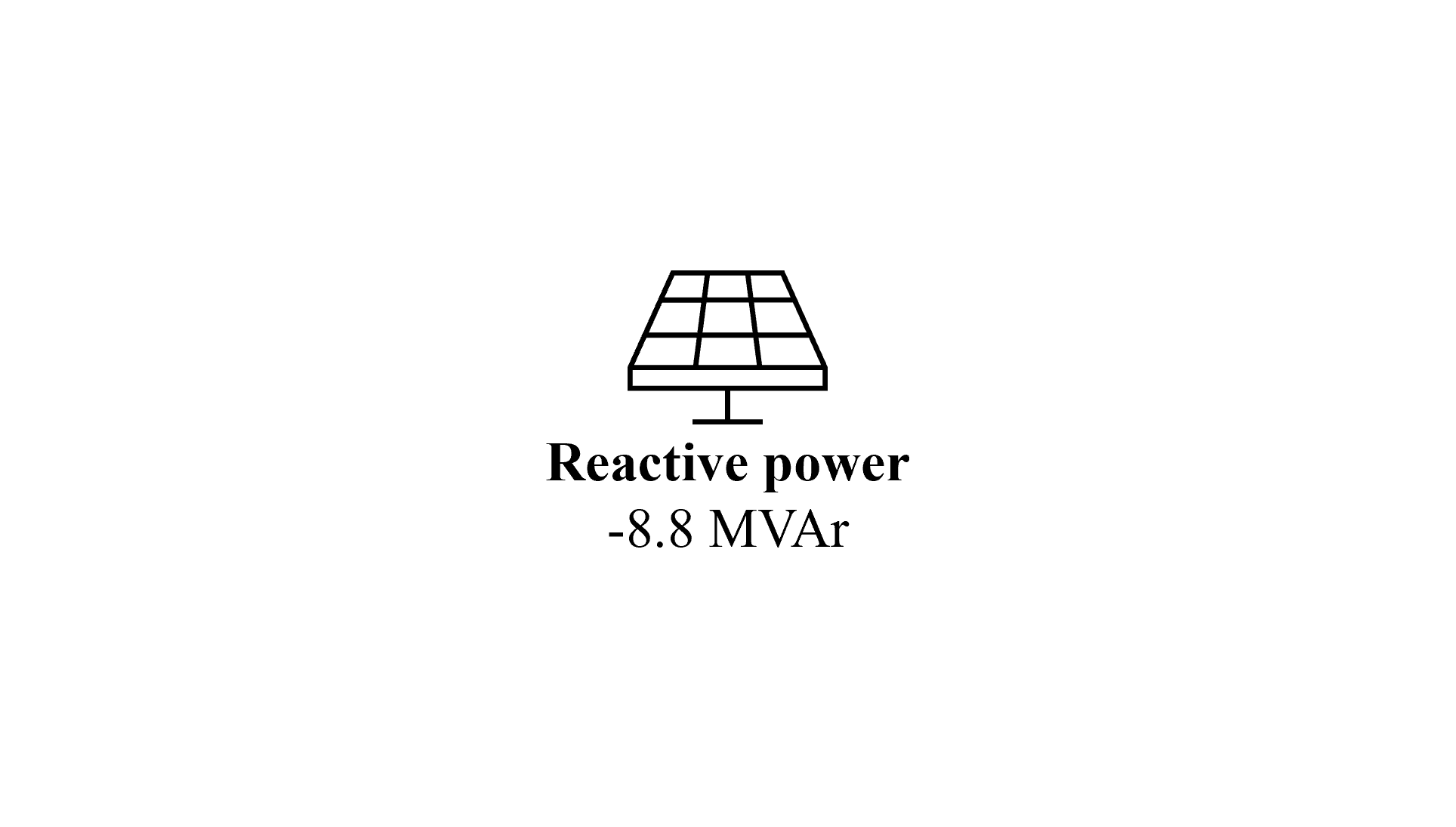}
         \caption{Reactive power photovoltaics}
         \label{fig:reac_pv}
        \end{subfigure}
     \hfill
     \begin{subfigure}[b]{0.45\textwidth}
         \centering
         \includegraphics[width=\textwidth]{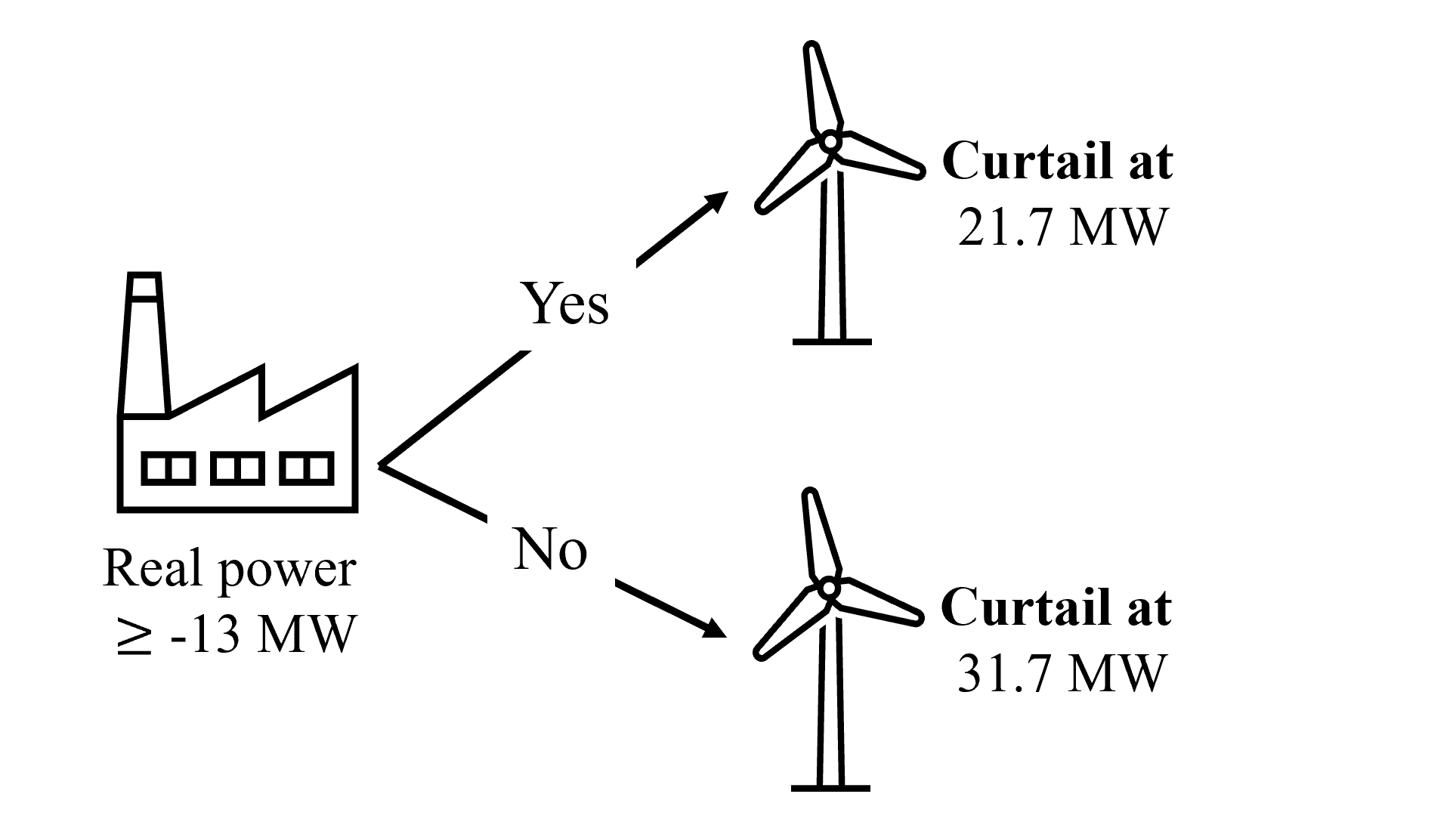}
         \caption{Curtailment wind-turbines}
         \label{fig:curt_wind}
     \end{subfigure}
     \hfill
     \begin{subfigure}[b]{0.45\textwidth}
         \centering
         \includegraphics[width=\textwidth]{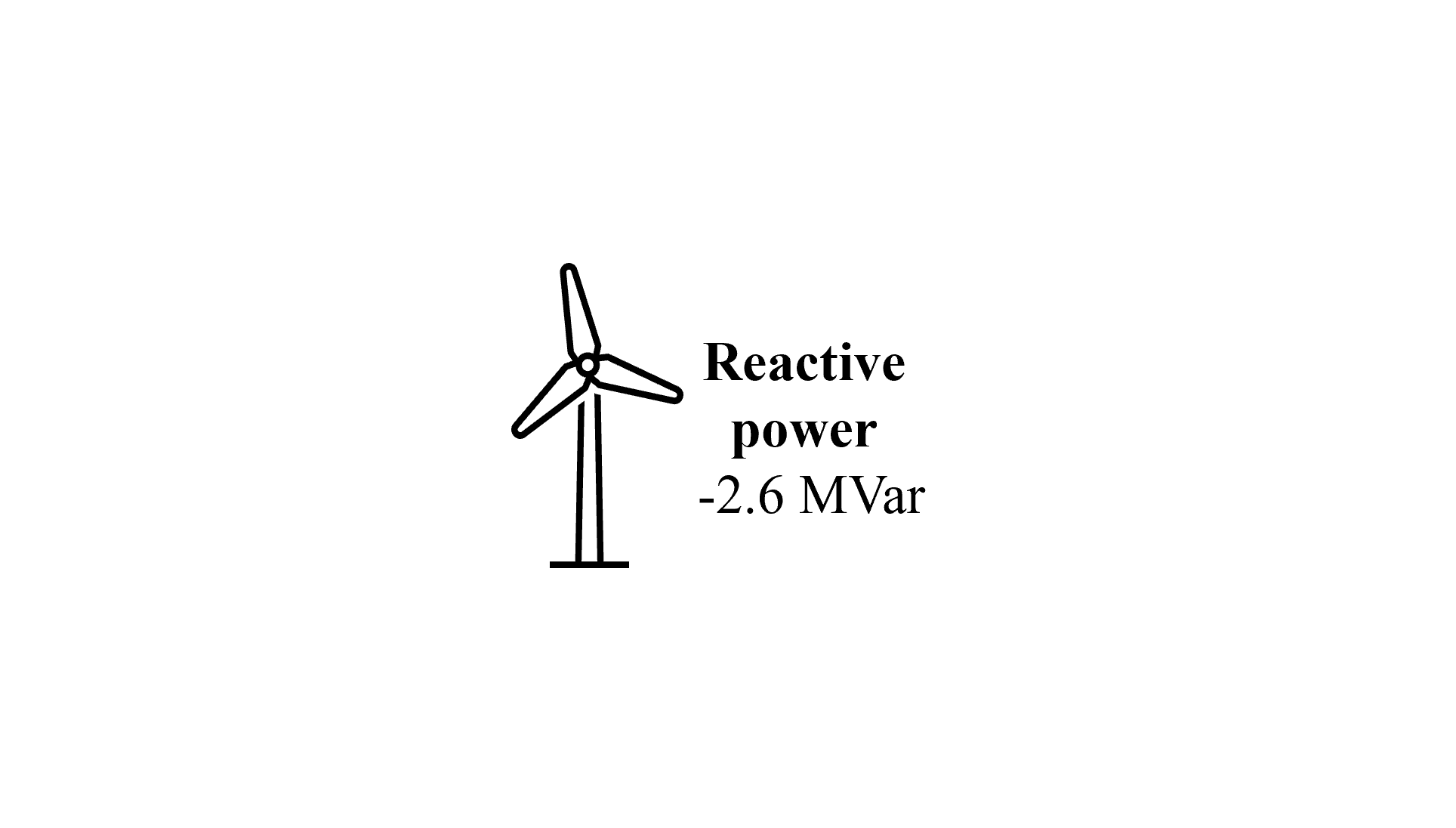}
         \caption{Reactive power wind-turbines}
         \label{fig:reac_wind}
     \end{subfigure}
     \hfill
     \begin{subfigure}[b]{0.45\textwidth}
         \centering
         \includegraphics[width=\textwidth]{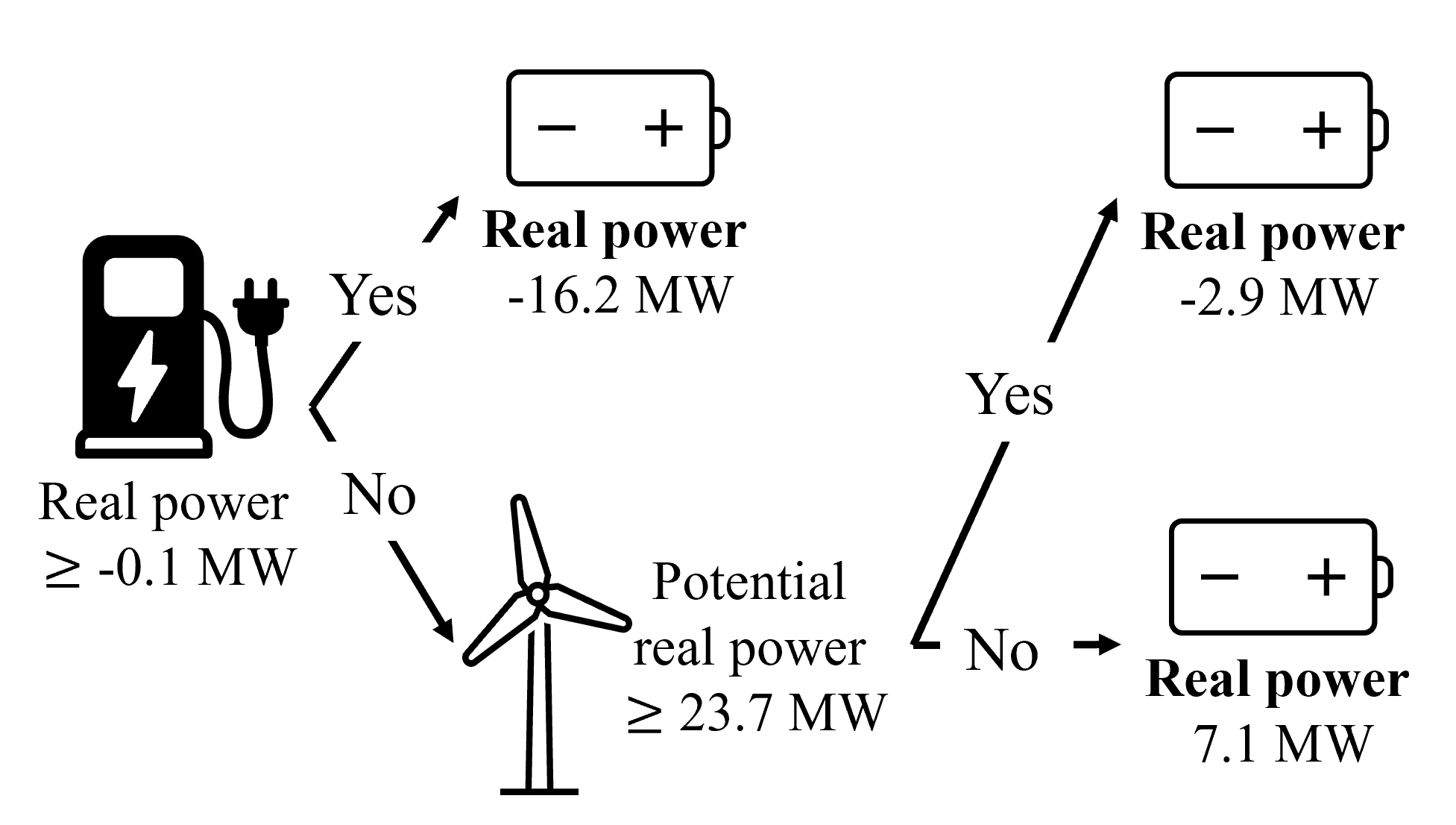}
         \caption{Real power large electrical storage}
         \label{fig:real_batt}
     \end{subfigure}
     \hfill
     \begin{subfigure}[b]{0.4\textwidth}
         \centering
         \includegraphics[width=\textwidth]{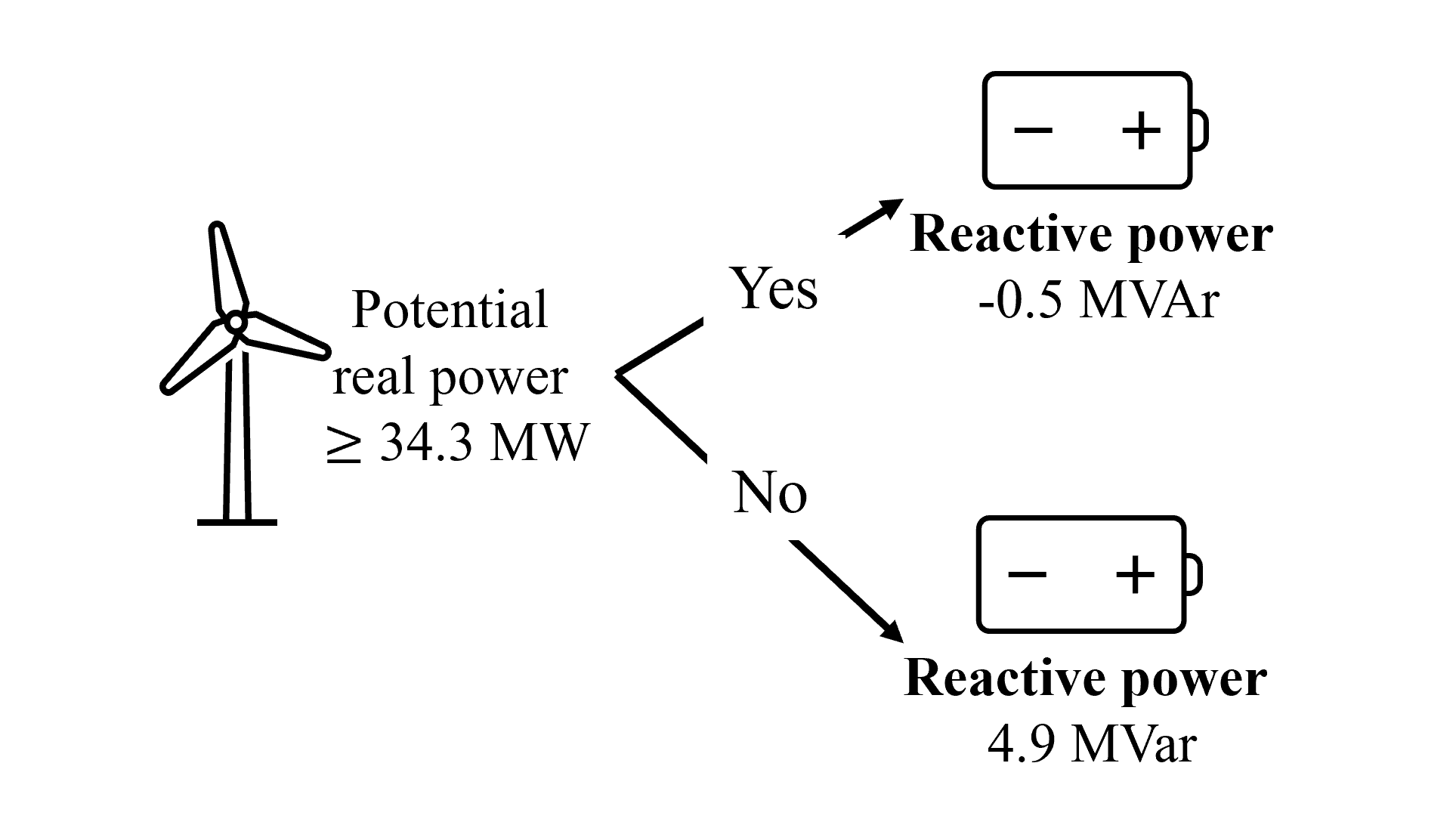}
         \caption{Reactive power large electrical storage}
         \label{fig:reac_batt}
     \end{subfigure}
     
        \caption{TreeC EMS which obtained the best score in Fig. \ref{fig:ANM_all_box}. Each tree controls the real power or reactive power of each controllable asset. To keep them simple for visualisation, nodes that were visited 10 times or less during evaluation were removed.}
        \label{fig:anm_trees}
\end{figure}
The trees that obtained the best performance shown in Fig. \ref{fig:anm_trees} provide a simple and interpretable EMS shown in \ref{fig:anm_trees} that obtains performances only slightly worse than an MPC based EMS with perfect forecast. 
The tree in Fig. \ref{fig:curt_pv} curtails PV at 21.2 MW when the power of the wind-turbines is low and otherwise curtails at 14.5 MW when the large electrical storage charges and at 6.2 MW when the large electrical storage discharges. 
The tree in Fig. \ref{fig:reac_pv} keeps the reactive power of the PV at -8.8 MVAr. 
The tree in Fig. \ref{fig:curt_wind} curtails wind-turbines at 31.7 MW when the industry load exceeds 13 MW and curtails at 21.7 MW otherwise. 
The tree in Fig. \ref{fig:reac_wind} keeps the reactive power of the wind-turbines at -2.6 MVAr. 
The tree in Fig. \ref{fig:real_batt} charges the large electrical storage at 16.2 MW when there is no demand from the electric vehicle charging garage otherwise charges at 2.9 MW when the wind-turbines can produce more than 23.7 MW and discharges at 7.1 MW when they can produce less. 
The tree in Fig. \ref{fig:reac_batt} sets the reactive power of the large electrical storage to -0.5 \deleted{
	\label{ch:mvar_typo}MVAr
	} MVAr when the wind-turbines can produce more than 34.3 MW and to 4.9 MVAr otherwise.

The interpretability of the trees gives interesting insight on the problem itself, for example in this case the reactive power are very easy to control efficiently since the trees obtained have a maximum of two different power settings for each assets (see Figs. \ref{fig:reac_pv}, \ref{fig:reac_wind} and \ref{fig:reac_batt}).

\subsection{BOPTEST Hydronic Heat Pump case}

\begin{figure}[H]
\centering
\includegraphics[width=0.85\linewidth]{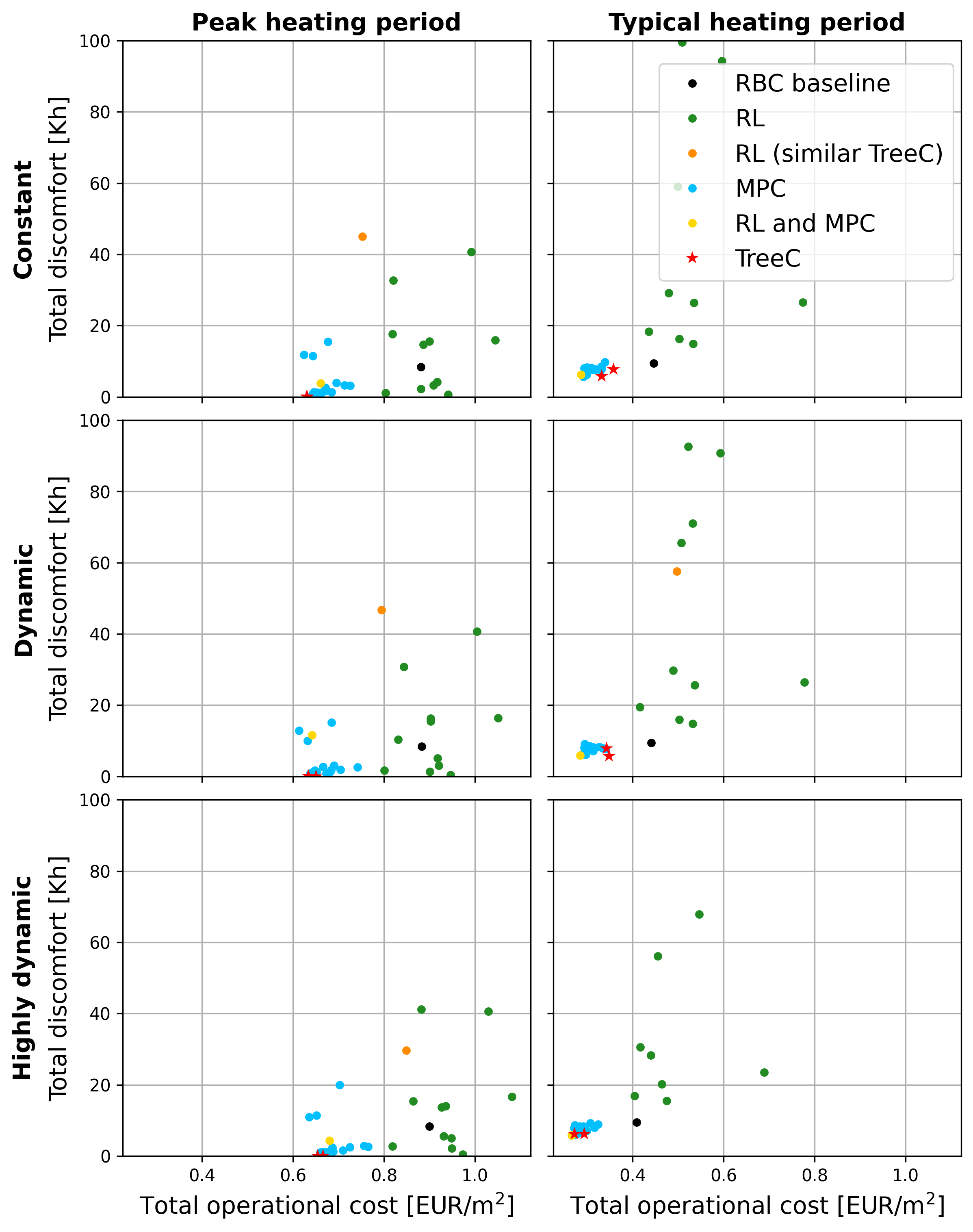}
\caption{BOPTEST results for the different pricing scenarios and validation time periods. The red stars are the results of the TreeC EMS\added{s}. The orange dot is the RL based EMS that had exactly the same inputs and actions as the TreeC EMS\added{s}. For MPC and RL based EMS\added{s}, the different dots use different combination of parameters, inputs and actions. 
Some RL based EMS\added{s} have scores with a worse \replaced{electricity}{operational} cost and total discomfort than the visualisation range selected here (see \cite{arroyo_comparison_2022} for full and detailed results). }
\label{fig:boptest_tree_res}
\end{figure}

In the second study case, running the evaluation of one individual simulation of two weeks lasted ~150 seconds. The time to update the CMA-ES optimisation model is negligible here. The trees generated for the results shown in \replaced{\label{ch:figure_typo}Fig. \ref{fig:boptest_tree_res}}{Fig. \ref{fig:ANM_all_box}} all ran for  12\,000 evaluation which amounts to ~20 days and 20 hours of training using a single thread and to ~6 hours and 15 minutes using the maximum number of 80 threads (population size of 16 $\times$ 5 parallel training). 

As shown in Fig. \ref{fig:boptest_tree_res} the TreeC EMS obtains similar performances to the best performing MPC based EMS for all price scenarios of the peak heating period and the highly dynamic price scenario of the typical heating period. 
For the dynamic and constant price scenarios of the typical heating period, the TreeC EMS performs similarly in terms of comfort and slightly worse in terms of cost than the MPC based EMS. 
In all scenarios the TreeC EMS performed better than the RBC baseline EMS and the RL EMS. 
This is also the case for the RL based EMS that has exactly the same inputs and actions as the TreeC EMS. 

\begin{table}[H]
	\centering
	\caption{\added{Average and standard deviation of the adjusted performances of the EMSs for the BOPTEST case.}}
	\begin{tabular}{|c|c|c|}
		
	\hline
	
	\added{EMS} & \added{Electricity cost [EUR/m\textsuperscript{2}]} & \added{Total discomfort [Kh]}  \\ \hline
	\added{RL and MPC} & \added{\textbf{0.018 $\pm$ 0.019}} & \added{3.42 $\pm$ 4.04} \\ \hline
	\added{TreeC} & \added{0.033 $\pm$ 0.021} & \added{\textbf{0.42 $\pm$ 0.77}} \\ \hline
	\added{MPC} & \added{0.037 $\pm$ 0.029} & \added{2.91 $\pm$ 3.67} \\ \hline
	\added{RBC} & \added{0.208 $\pm$ 0.056} & \added{6.05 $\pm$ 2.34} \\ \hline	
	\added{RL} & \added{0.266 $\pm$ 0.085} & \added{24.41 $\pm$ 23.47} \\ \hline

	\end{tabular}
	
	\label{tab:boptest_table_results}
\end{table}

\added{Tab. \ref{tab:boptest_table_results} shows the adjusted and averaged electricity cost and total discomfort obtained by each EMS category in Fig. \ref{fig:boptest_tree_res}.
The electricity cost and total discomfort are adjusted in this table by subtracting the value of the best performing EMS for each metric on each of the 6 different scenarios (different price and heating periods).
This shows how far on average each EMS category is from the best performing EMS for each metric over the 6 scenarios.
}

\begin{figure}[H]
\centering
\includegraphics[width=0.5\linewidth]{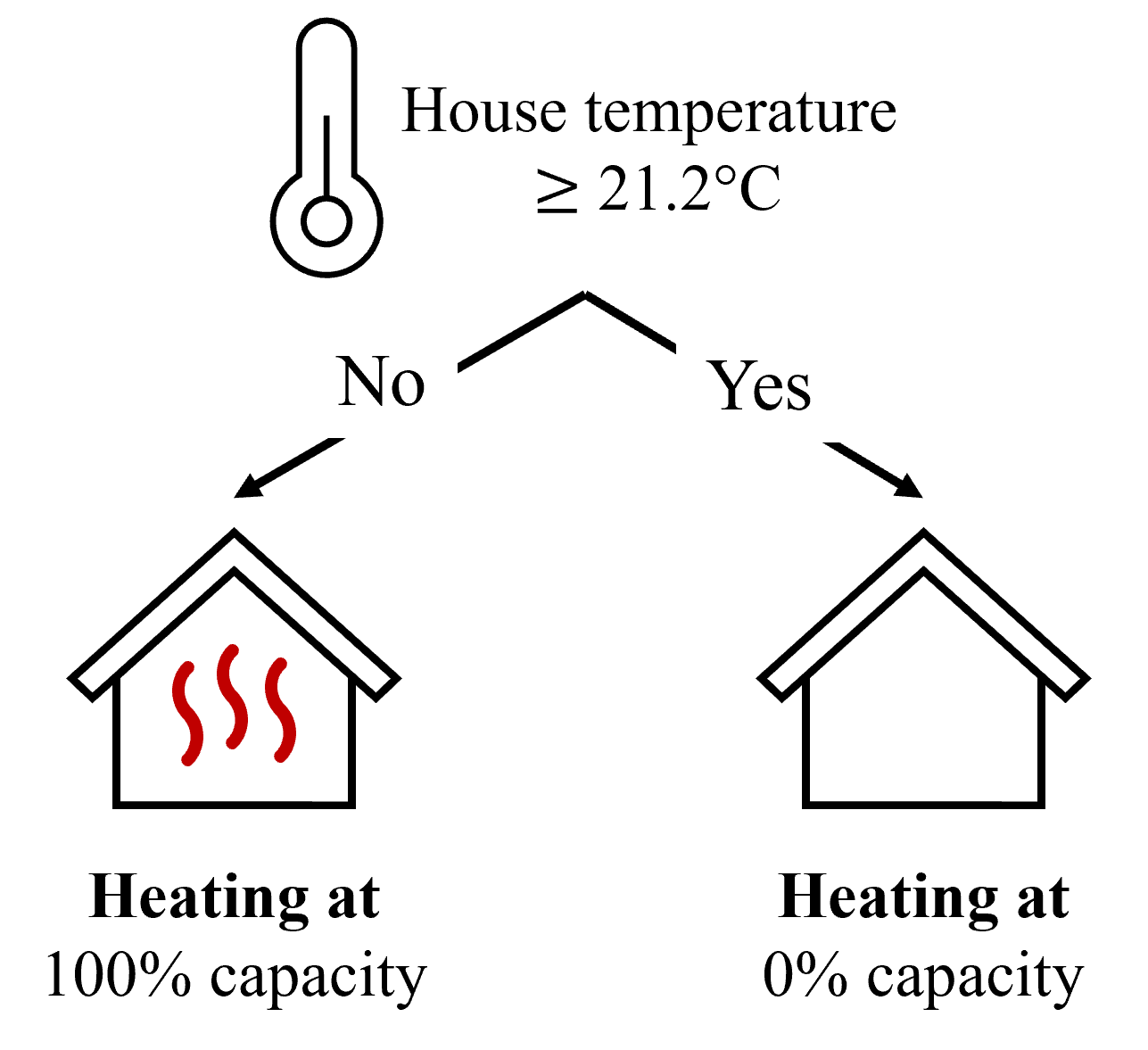}
\caption{TreeC EMS with best score obtained on the BOPTEST case for the constant price and peak heating period scenario. }
\label{fig:boptest_tree}
\end{figure}
\begin{figure}[H]
\centering
\includegraphics[width=0.9\linewidth]{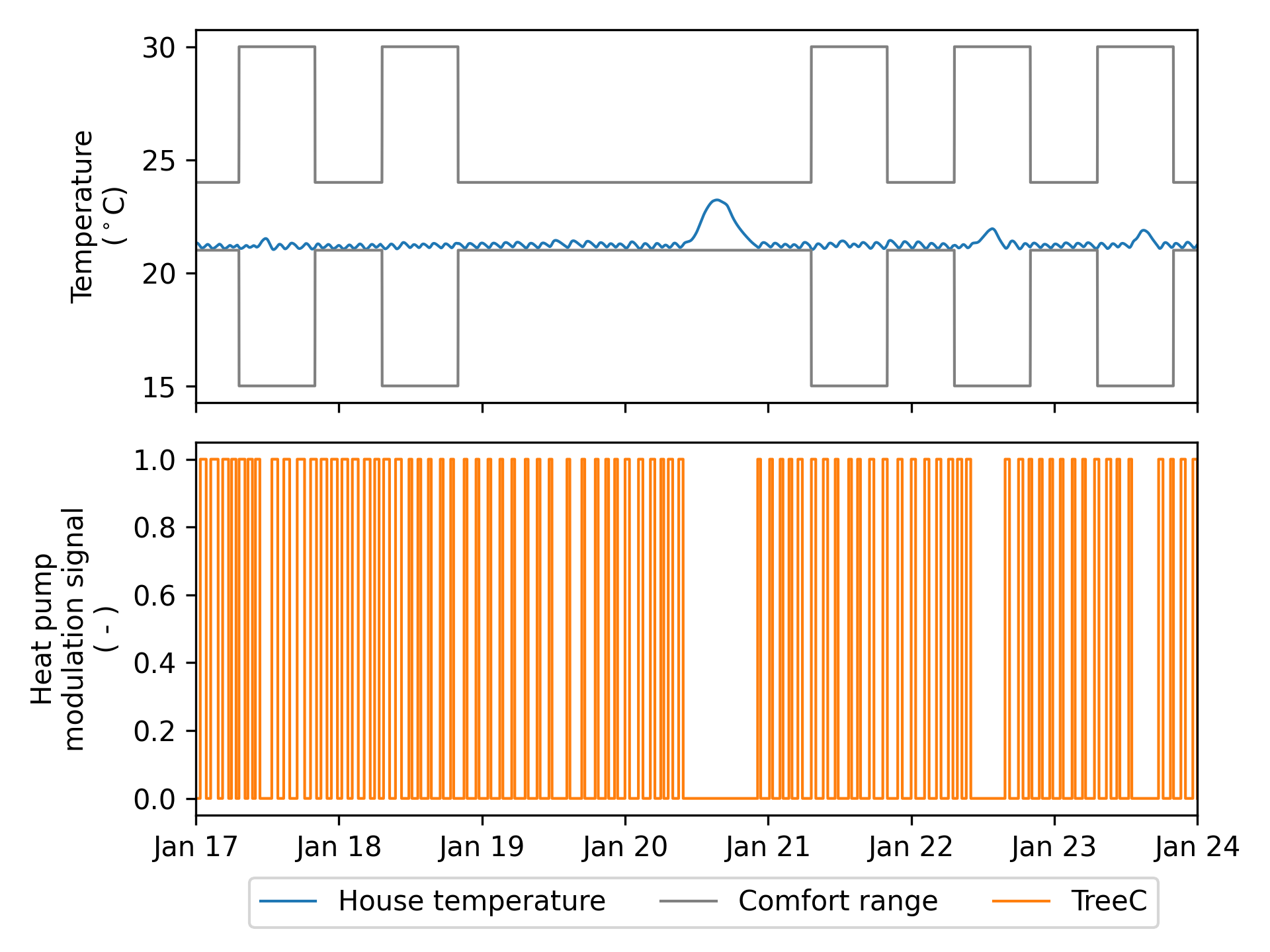}
\caption{TreeC EMS presented in Fig. \ref{fig:boptest_tree} controlling the heat pump for the first week of the peak heating validation period. The top plot shows that the indoor temperature always stays just above the lower bound of the comfort range during occupied hours. The increase in indoor temperature particularly noticeable on the \nth{20} of January is mainly caused by solar irradiation.}
\label{fig:boptest_vis}
\end{figure}

The trees obtained in training were all very similar to the one shown in Fig. \ref{fig:boptest_tree}. It is a  very simple tree that maintains the room temperature around 21.2\textdegree C which is just above the 21\textdegree C lower bound of the comfort range when the house is occupied. This behaviour ensures that the house temperature is always above the lower bound of the comfort range while keeping a relatively low temperature and therefore not having to buy too much electricity (see Fig \ref{fig:boptest_vis}). The MPC behaviour is similar during occupied time periods but sometimes goes close to 20\textdegree C during unoccupied time periods when calculated as profitable. The TreeC EMS results show that going close to 20\textdegree C during the unoccupied time periods does not improve the score compare to simply keeping the temperature just above 21\textdegree C.

In this case TreeC found an EMS that is very performant and much more practical to implement than an MPC or RL based EMS.

\section{Discussion}

\subsection{Advantages and disadvantages of different EMS methods}

Based on the studies done in this paper, we identify advantages and disadvantages of different EMS methods and summarised them in Tab. \ref{tab:comp}. 

\begin{table}[H]
\fontsize{8}{10}\selectfont
\caption{Relevant comparison criteria for the TreeC, MPC, RL and RBC methods.}
\label{tab:comp}
\resizebox{1.0\textwidth}{!}{%
\begin{tblr}{
colspec={|l|l|l|l|l|},
cell{2}{2} = {green},
cell{2}{3} = {green},
cell{2}{4} = {green},
cell{2}{5} = {},
cell{3}{2} = {green},
cell{3}{3} = {green},
cell{3}{4} = {red},
cell{3}{5} = {orange},
cell{4}{2} = {green},
cell{4}{3} = {green},
cell{4}{4} = {red},
cell{4}{5} = {orange},
cell{5}{2} = {red},
cell{5}{3} = {green},
cell{5}{4} = {red},
cell{5}{5} = {green},
cell{6}{2} = {red},
cell{6}{3} = {green},
cell{6}{4} = {green},
cell{6}{5} = {orange},
cell{7}{2} = {teal7},
cell{7}{3} = {red},
cell{7}{4} = {green},
cell{7}{5} = {red},
cell{8}{2} = {green},
cell{8}{3} = {red},
cell{8}{4} = {green},
cell{8}{5} = {green},
cell{9}{2} = {green},
cell{9}{3} = {teal7},
cell{9}{4} = {red},
cell{9}{5} = {green},
cell{10}{2} = {green},
cell{10}{3} = {red},
cell{10}{4} = {green},
cell{10}{5} = {green},
cell{11}{2} = {red},
cell{11}{3} = {orange},
cell{11}{4} = {orange},
cell{11}{5} = {green},
}
\hline
                    & TreeC                 & MPC         & RL                 & RBC   \\ \hline
ANM performance                                  & \added{19.2}                  & \added{14.4}        & \added{16.2}               & None        \\ \hline
BOPTEST performance                              & \added{0.03 EUR/m\textsuperscript{2}}                  & \added{0.04 EUR/m\textsuperscript{2}}        & \added{0.27 EUR/m\textsuperscript{2}}                & \added{0.21 EUR/m\textsuperscript{2}}     \\
& \added{0.42 Kh}                  & \added{2.91 Kh}        & \added{24.41 Kh}               & \added{6.05 Kh}     \\ \hline
Common real implementations                             & No                    & Yes         & No                 & Yes         \\ \hline
Many possible output values         & No & Yes         & Yes                & Moderate         \\ \hline
\added{Adapts model to problem}
                                  & In simulation               & No          & In real case & No          \\ \hline
Scales well with case complexity         & Yes                  & No    & Yes           & Yes        \\ \hline
Interpretability of model                       & Very good             & Good        & Bad                & Very good   \\ \hline
Relies on good forecasting & No                    & Yes         & No                 & No          \\ \hline

Training time                 & Very high             & Forecast & High               & No training \\ \hline

\end{tblr}}

\end{table}

The outstanding advantages of the MPC method are that the performance is good and has proven itself in several real use cases. 
The disadvantages are that the method is deterministic and does not have a learning behaviour. 
\replaced{MPC}{Therefore MPC does not adapt to the case it is implemented in, it} relies on good forecasting of the system's various inputs and \replaced{needs to model the system it controls}{the system it controls must be modeled} using mathematical programming. 
This last point has the disadvantage that one must understand the system very well in order to be able to model it, and in almost all cases model simplifications are necessary in order to be able to solve the mathematical programming problem in the required time. 
\added{MPC does not automatically adapt the control model to the problem by for example changing model parameters when the forecast is not accurate.}
The interpretability is good (i.e. the MPC controls the \replaced{system}{EMS} perfectly assuming all given models and inputs are accurate) but in most cases a lot of inputs and model constraints need to be taken into account to completely interpret the decision process of the MPC. 

On the other hand, the TreeC and RL methods both have the advantage of \replaced{learning and adapting their control model}{being able to learn and adapt} to the controlled system. 

The RL method is capable of learning online meaning that RL can adapt the EMS while it is controlling the system. This is a real advantage as it would receive a very accurate feedback from the system and not have to rely on a less accurate simulation of the system. Nonetheless many training steps, corresponding to multiple years, of bad performance are needed for the RL model to obtain a good performance \cite{ceusters_model-predictive_2021}. Hence, either training in simulation or a transfer learning method would still be necessary to implement this in a real case. 

The RL method uses a neural network as control model. This allows the model to have very complex behaviours sometimes necessary to solve certain problems. On the other hand it is a black box model which is not interpretable and therefore could lead to unexpected behaviours which is not desirable for an EMS. 

For the two studied cases, RL fails to obtain consistently a good performance. In the ANM6Easy case, SAC obtained a good performance but not PPO. In the BOPTEST case, all RL algorithm performed poorly even compared to the simple RBC baseline.  The trade-off of RL for using a model that is not interpretable should be that it is capable of solving simple and complex cases equally but there is no clear methodology in literature yet shown to obtain consistently good results over different benchmark cases. 

The main advantage of the TreeC method is the interpretability of its models as shown in Fig. \ref{fig:anm_trees} and \ref{fig:boptest_tree}. This allows total confidence in the EMS when implementing it in a real case, the possibility to recognise some possible wrong behaviour that can be caused by the difference between the simulation and real case and makes it easier to transfer the knowledge gained on how to control this system to another similar system.

Another advantage is that the method does not require much customisation other than the necessary ones for any machine learning problem (i.e. choosing good model inputs and outputs, which training steps to use and for how long should the training process run). The CMA-ES algorithm performs very well with default hyper-parameters and the number of nodes of the tree can be defined to make the tree more or less interpretable. TreeC obtained consistently good performance over both the ANM6Easy and BOPTEST cases changing only the part of the methodology specific to the problem (i.e. inputs, outputs, training steps and training length). 

One main disadvantage is that it cannot learn online and therefore requires an accurate simulation of the system it needs to control. This simulator also needs to be relatively fast as TreeC needs to simulate many steps to find a good solution. The CMA-ES algorithm allows to parallelise the evaluation of individuals within a generation but a fast simulator is still important to be able to do the individual's evaluation over many steps and the training over many generations. 

Another disadvantage is that the complexity of the behaviour of the model is limited by the size of the tree, one tree has a maximum number of output values limited to the number of leafs of the tree and the tree should remain small to keep it interpretable. For certain more complex problems this can definitely be an issue that needs to be addressed.

\subsection{Improvements}

TreeC can be improved and further validated in the following ways:
\begin{itemize}
    \item To address the issue of not being able to have a complex behaviour, the leafs could be a RBC or MPC variations instead of single output values. This would allow to tune RBC and MPC methods based on the case they are implemented in.
    \item CMA-ES could be replaced by a metaheuristic algorithm that optimises directly in the mixed continuous and discrete search space instead of optimising in the continuous space and discretising variables when necessary.  
    \item The time periods used to perform the training were chosen in a rather arbitrary manner, a further study is need to define what is a good training period and how to improve an already trained model when new data becomes available. 
    \item The method should be evaluated on more complex cases. This would contribute to understand which are the limits for the applicability of the proposed method. 
    \item The method should be implemented and evaluated in a real case to measure the importance of the drop in performance between simulation and reality.

\end{itemize}

\section{Conclusion}

This paper presents TreeC, a new methodology to generate EMS modeled as decision trees. 
The method \replaced{makes}{aims at making} the learned EMS model interpretable while also \replaced{optimising}{obtaining good} performances. 
TreeC was compared with \deleted{different} MPC and RL methods in the ANM6Easy and BOPTEST \added{Hydronic Heat Pump} cases. 
\replaced{
\label{ch:conclusion_results}	
In the ANM6Easy case, TreeC achieves an average energy loss and constraint violation score of 19.2, which is close to MPC with perfect forecast and SAC with no outliers that achieve scores of 14.4 and 16.2 respectively. 
All three methods control the ANM6Easy case well especially when compared to the random EMS, which obtains an average score of 12\,875. 
In the BOPTEST case, TreeC performs similarly to MPC with perfect forecast on the adjusted and averaged electricity cost and total discomfort (0.033 EUR/m\textsuperscript{2} and 0.42 Kh for TreeC compared to 0.037 EUR/m\textsuperscript{2} and 2.91 kH for MPC), while outperforming RBC (0.208 EUR/m\textsuperscript{2} and 6.05 Kh) and RL (0.266 EUR/m\textsuperscript{2} and 24.41 Kh). 
In both cases, the decision trees obtained by TreeC are simple and interpretable.
The interpretable nature of the decision trees give interesting insights on the case itself.
In the ANM6Easy case, the reactive power of the photovoltaics and wind-turbines are controlled efficiently by simply setting them to a constant value.
In the BOPTEST case, the system is controlled efficiently by constantly keeping the indoor temperature just above the lower bound of the comfort range for occupied hours.
}
{It was shown to consistently perform only slightly worse than MPC methods with perfect forecast in the ANM6Easy and BOPTEST cases and even performing among the best MPCs in 4 out of 6 scenarios of the BOPTEST case. 
Compared to RL methods implemented in the ANM6Easy case, TreeC performed better with the same inputs and outputs as the PPO based method and slightly worse than the SAC based method but only when excluding bad performing outliers. 
In the BOPTEST case, TreeC consistently outperformed the different RL methods using very simple decision trees.}

TreeC generates interpretable EMSs and obtains consistently good performances over the two evaluated cases without needing to \replaced{make case-specific adjustments to}{change} the method. 
Future works should address the limited complexity of the possible behaviour of the EMS and evaluate it in more difficult cases as well as implement the method in a real life case.


\section*{CRediT authorship contribution statement}

\textbf{Julian Ruddick:} Conceptualization, Methodology, Software, Validation,  Formal analysis, Investigation,  Data Curation, Writing - Original Draft, Writing - Review \& Editing, Visualization. \textbf{Luis Ramirez Camargo:} Conceptualization, Writing - Review \& Editing, Supervision. \textbf{Muhammad Andy Putratama:} Writing - Review \& Editing, Visualization, Supervision. \textbf{Maarten Messagie:} Supervision, Funding acquisition. \textbf{Thierry Coosemans:} Supervision, Funding acquisition.

\section*{Declaration of competing interest}
The authors declare that they have no known competing financial interests or personal relationships that could have appeared to influence the work reported in this paper.

\section*{Data availability}

The results and code necessary to reproduce the results are available on the following public GitHub repository: \url{https://github.com/EVERGi/treec-paper-results}.

\section*{Acknowledgement}

This work has been supported by the ICON project OPTIMESH (FLUX50 ICON Project Collaboration Agreement - HBC.2021.0395) funded by VLAIO and by the ECOFLEX project funded by FOD Economie, K.M.O., Middenstand en Energie.

We kindly thank Javier Arroyo for providing the plotting script of Fig. \ref{fig:boptest_tree_res} and sharing useful additional information on his paper \cite{arroyo_comparison_2022}.

\bibliographystyle{elsarticle-num-names} 
\bibliography{cas-refs.bib}

\end{document}